**SmartAlert: Implementing Machine Learning-Driven Clinical Decision Support for Inpatient Lab Utilization Reduction**

April S. Liang, MD*[1]; Fatemeh Amrollahi, PhD*[2]; Yixing Jiang, BE*[2]; Conor K. Corbin, PhD[3]; Grace Y.E. Kim, MS[4]; David Mui, MD, MBA[5]; Trevor Crowell, BA[6]; Aakash Acharya[7]; Sreedevi Mony[7]; Soumya Punnathanam[7]; Jack McKeown[7]; Margaret Smith, MBA[6,8]; Steven Lin, MD[6,8]; Arnold Milstein, MD, MPH[8,9]; Kevin Schulman, MD, MBA[1,9]; Jason Hom, MD[1]; Michael A. Pfeffer, MD, FACP[1,7]; Tho D. Pham, MD[10]; David Svec, MD, MBA[1,11]; Weihan Chu, MD[1,11]; Lisa Shieh, MD, PhD[1]; Christopher Sharp, MD[8]; Stephen P. Ma, MD, PhD**[1]; Jonathan H. Chen, MD, PhD**[1,2,9,12]

*Drs. Liang, Amrollahi, and Yixing Jiang contributed equally to this manuscript. **Drs. Ma and Chen contributed equally to this manuscript.

[1]Division of Hospital Medicine, Department of Medicine, Stanford University School of Medicine, Palo Alto, CA.
[2]Department of Biomedical Data Science, Stanford University, Palo Alto, CA.
[3]SmarterDx, New York, NY
[4]The Johns Hopkins University School of Medicine, Baltimore, MD
[5]Department of Medicine, Stanford University School of Medicine, Palo Alto, CA.
[6]Stanford Healthcare AI Applied Research Team, Stanford University School of Medicine, Palo Alto, CA.
[7]Technology and Digital Solutions, Stanford Health Care, Palo Alto, CA.
[8]Division of Primary Care and Population Health, Department of Medicine, Stanford University School of Medicine, Palo Alto, CA.
[9]Clinical Excellence Research Center, Stanford University, Palo Alto, CA
[10]Department of Pathology, Stanford University School of Medicine, Palo Alto, CA
[11]Stanford Health Care Tri-Valley, Pleasanton, CA
[12]Stanford Center for Biomedical Informatics Research, Stanford University, Palo Alto, CA

Corresponding Author:

April S. Liang
Stanford University Medical Center Hospital Medicine
300 Pasteur Dr.
E Pavilion Fl 3 MC 5158
Palo Alto, CA 94304
asliang@stanford.edu

Word Count: 2044

**Abstract**

Repetitive laboratory testing unlikely to yield clinically useful information is a common practice that burdens patients and increases healthcare costs. Education and feedback interventions have limited success, while general test ordering restrictions and electronic alerts impede appropriate clinical care. We introduce and evaluate SmartAlert, a machine learning (ML)-driven clinical decision support (CDS) system integrated into the electronic health record that predicts stable laboratory results to reduce unnecessary repeat testing. This case study describes the implementation process, challenges, and lessons learned from deploying SmartAlert targeting complete blood count (CBC) utilization in a randomized controlled pilot across 9270 admissions in eight acute care units across two hospitals between August 15, 2024, and March 15, 2025. Results show significant decrease in number of CBC results within 52 hours of SmartAlert display (1.54 vs 1.82, $p < 0.01$) without adverse effect on secondary safety outcomes, representing a 15% relative reduction in repetitive testing. Implementation lessons learned include interpretation of probabilistic model predictions in clinical contexts, stakeholder engagement to define acceptable model behavior, governance processes for deploying a complex model in a clinical environment, user interface design considerations, alignment with clinical operational priorities, and the value of qualitative feedback from end users. In conclusion, a machine learning-driven CDS system backed by a deliberate implementation and governance process can provide precision guidance on inpatient laboratory testing to safely reduce unnecessary repetitive testing.

**Description (1-2 sentences)**

A SmartAlert – a machine learning-driven clinical decision support system integrated into the electronic health record – targeting inpatient complete blood count utilization was deployed and associated with reduced repetitive testing. Intervention success depended on stakeholder alignment, end user engagement, deliberate implementation, and systematic governance.

## Introduction

More than 7 billion clinical lab tests are performed annually in the United States[1], and 20-50% of inpatient lab tests are medically unnecessary[2–4], contributing to patient harms including worsening anemia[5–7], sleep disruption[8], and increased healthcare costs[9]. The Society of Hospital Medicine identifies routine complete blood counts (CBC) as a wasteful practice in their "Choosing Wisely" list[10]. Existing mitigation strategies, including clinician education, audit and feedback, financial incentives, and broad ordering restrictions, have limited precision and efficacy[11,12]. Clinician cognitive overload and concerns for patient safety pose persistent barriers to reducing lab overutilization[13–15].

In late 2021, global supply chain disruptions related to the COVID-19 pandemic resulted in critical shortages of blood collection tubes and overt nationwide rationing[16], highlighting these issues. Our institution implemented ordering restrictions that limited the duration of standing lab draws to 72 hours. While successful in reducing lab testing, this intervention blocked repeat testing even for patients undergoing dynamic clinical change. A more precise approach to safely reduce unnecessary laboratory testing is needed.

Advances in machine learning (ML) provide a promising avenue for addressing this challenge by offering clinicians patient-specific identification of low-value lab testing[14,17], but real-world use remains limited by technical, operational, and clinical barriers. We introduce SmartAlert, a novel ML-driven real-time clinical decision support (CDS) tool integrated into the electronic health record (EHR). We describe the implementation process, challenges, and lessons learned from our randomized deployment of a SmartAlert targeting repetitive inpatient CBCs and its impact on CBC testing rates and secondary safety outcomes.

## Methods

### Study Setting

In response to global supply shortages, a multidisciplinary team at our institution was tasked in 2022 with developing methods to provide targeted recommendations to clinicians on unnecessary lab testing. The CBC utilization SmartAlert was deployed at two facilities within our health system: an academic tertiary care hospital and an affiliated community hospital. Both facilities shared a single EHR instance (Epic Systems, Verona, WI) and at baseline required clinician review and renewal of standing inpatient laboratory tests every 72 hours.

### SmartAlert System Architecture

The SmartAlert architecture is built upon the DEPLOYR framework,[18] which uses standard EHR components to trigger custom ML predictions using real-time data (Figure 1a):

1. Prediction Trigger: Using standard EHR functions, predefined user actions – such as ordering a lab with daily or higher frequency – can be used to trigger a RESTful (Representational State Transfer) HTTP (Hypertext Transfer Protocol) web request to an Azure (Microsoft Corporation, Redmond, WA) serverless function for a custom prediction calculation. Given that data retrieval could take several minutes, we opted to systematically trigger predictions in advance for all admitted patients at regular 6-hour intervals.

2. Data Retrieval: Relevant patient data, including demographics, lab results, vitals, medications, and diagnoses, are extracted using standard FHIR (Fast Healthcare Interoperability Resources) application programming interfaces (APIs) at the time of inference.

3. Prediction Module: The lab utilization SmartAlert uses a probabilistic regression model to predict the likelihood of lab test stability based on predefined thresholds, as previously described in Jiang et al.[17] The predictions are stored in custom EHR flowsheets.

4. Alert Display: At the time that a clinician attempts to order repetitive lab tests, a classic CDS alert (Epic Our Practice Advisory, OPA) is triggered within the EHR based on the stored predictions.

The CDS alert's user interface and workflow criteria were designed according to the capabilities of our EHR and with clinician user input, as described in 'User Co-Design and Silent Prospective Model Performance' (Figures 1b-c). The alert interface provides probability of lab stability, recent results, and a link for more information. Key workflow decisions included 1) displaying the alert only between 7am and 6pm to spare overnight coverage clinicians and 2) offering reduced lab test frequencies as an alternative to cancellation.

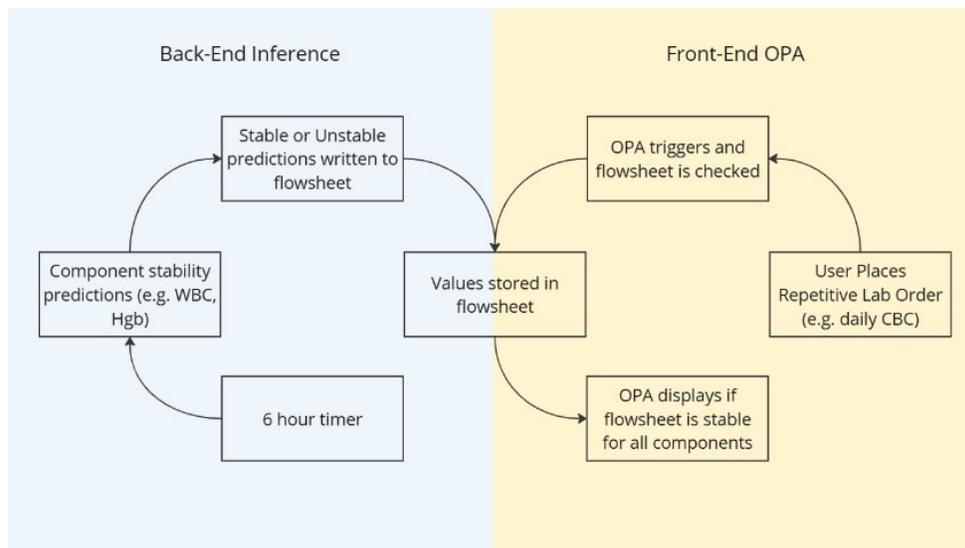

**Figure 1a: SmartAlert system architecture.** The DEPLOYR framework allows for SmartAlerts to initiate real-time patient data extraction, perform custom prediction calculation, and write back or trigger an intervention in the EHR. The process begins in response to a system generated web request - in this case a 6-hour timer - and ends with stability prediction being stored in a custom flowsheet. Classic CDS alerts (OPA Our Practice Advisory) then display in response to clinicians ordering repetitive labs.

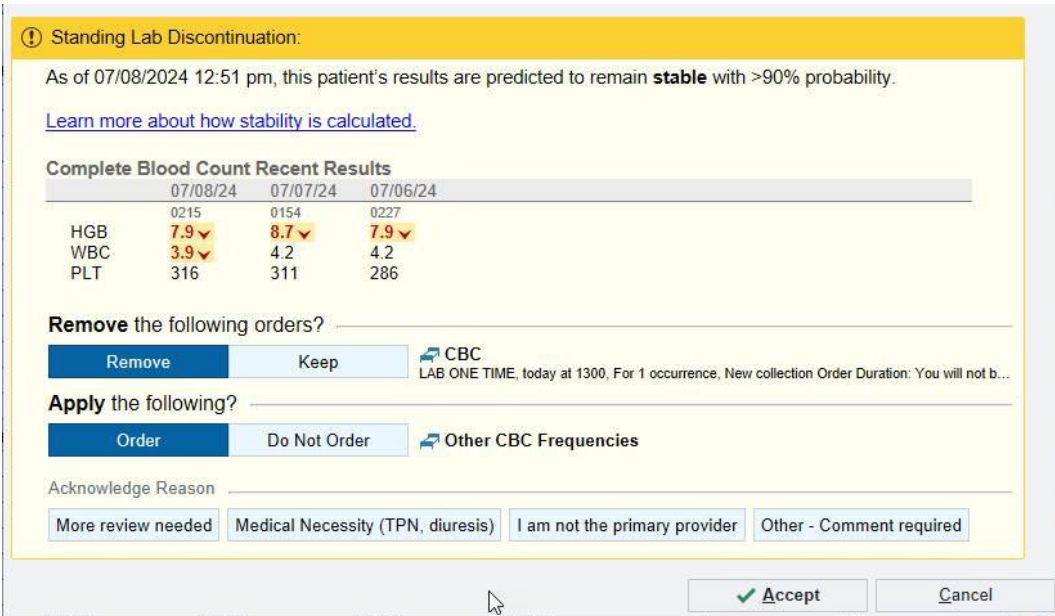

**Figure 1b: Clinician facing SmartAlert user interface.** If a clinician attempts to order further repetitive CBCs that are predicted to be stable, they are presented with this OPA interface to indicate the personalized prediction of >90% stability, a link to supporting information, table of recent lab result values, and multiple interaction options.

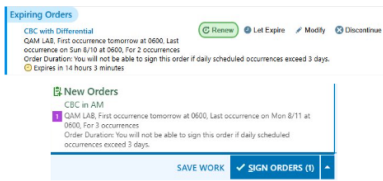 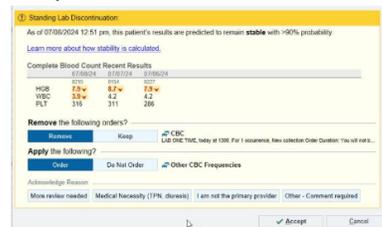 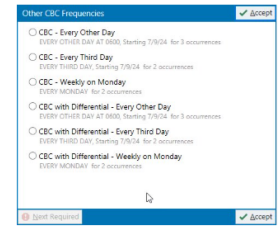

**Figure 1c: Clinician workflow.** On encountering the CBC utilization SmartAlert, the ordering clinician is given the option to acknowledge the SmartAlert to continue the repetitive lab test order, cancel the order, or reduce the frequency of the lab order (e.g. every other day instead of daily).

## Governance

### Retrospective Model Validation

We employed a multi-phase evaluation and governance process. During model development, performance on retrospective test data drawn from 7 years of admissions in our health system was tuned to achieve 90% precision (positive predictive value, PPV) previously identified by clinicians as meaningful[14] with recall (sensitivity) for white blood cell count, hemoglobin, and platelet count of 29.3%, 59.5%, and 100%, respectively[17].

### Initial Concept Approval

We presented the CBC utilization SmartAlert concept and retrospective validation to our institution's CDS Committee (Figure 2). This committee reviews all EHR-based CDS proposals to ensure they are meaningful and effective. Our team described the lab overutilization/blood tube shortage challenge, explained why an interruptive OPA was the appropriate intervention, and outlined measures of alert success. The Committee approved the concept, allowing us to proceed to detailed alert content and workflow design.

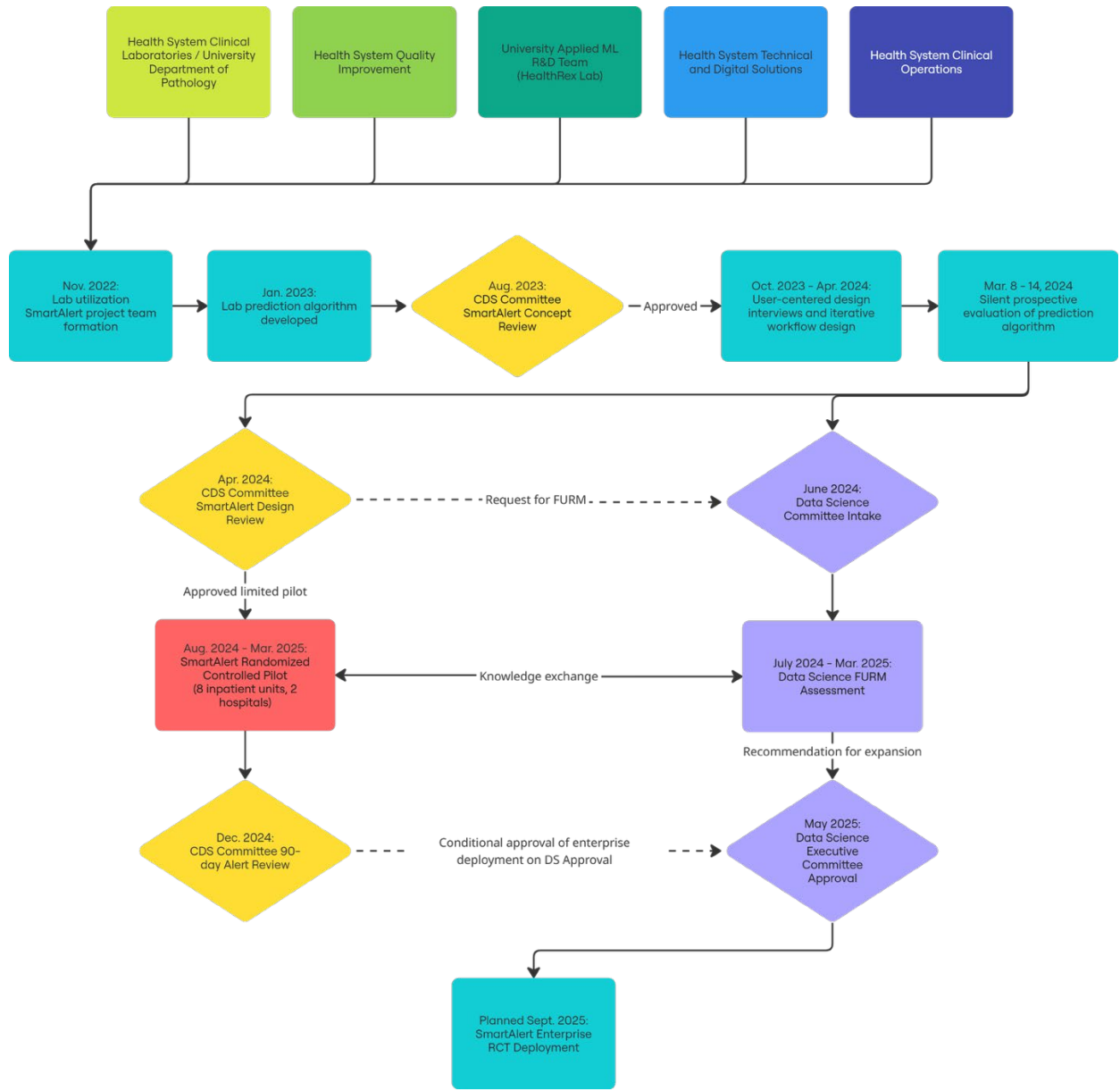

**Figure 2: CBC utilization SmartAlert development and governance process.** To implement the CBC utilization SmartAlert, the multidisciplinary team (turquoise) underwent a 30-month process of development, evaluation, and governance, including a 7-month randomized controlled pilot (red) described in this study, and involving the CDS Committee (yellow) and the Data Science teams (purple). ML, machine learning; R&D, research and development; CDS, clinical decision support; FURM, Fair, Useful, and Reliable AI Model; DS, data science.

**User Co-Design and Silent Prospective Model Performance**

To ensure usability and clinician trust, we conducted semi-structured interviews (Supplemental S1) prior to implementation (Figure 2). Trainees, attending physicians, and advanced practice providers with inpatient work experience from all specialties were invited to participate on a voluntary basis using public mailing lists, with additional convenience and purposive sampling performed through targeted outreach. 18 clinicians from Internal Medicine/Hospital Medicine, Neurology, Critical Care, General Surgery, Orthopedic Surgery, Hematology/Oncology, Urology, and Otolaryngology participated.

While earlier iterations of model development aimed to predict whether a lab result would be "normal," clinician stakeholders indicated that predicting "clinical stability" of a test was more meaningful[14]. Using the 18 participants' definitions of "clinically stable" lab values, we determined the stability thresholds for components of the CBC using absolute minima/maxima and acceptable directional relative changes (Supplemental S2). Using these thresholds, the CBC utilization SmartAlert underwent *silent* prospective evaluation between March 8 and March 14, 2024, operating in real time without triggering OPA displays. This confirmed that model PPV remained near the goal of 90% (e.g. 88.1%, 95.4%, and 93% for white blood cell count, hemoglobin, and platelet count, respectively).

Qualitative feedback from clinician interviews was inductively coded with consensus. These revealed favorable attitudes toward the use of ML in the alert with high trust and positive sentiment (Figure 3a). Many participants indicated that knowing the model's PPV was most helpful, while being ambivalent on the need to disclose the role of ML (Figure 3b-d). Additional feedback was iteratively incorporated into the design (Supplemental S4).

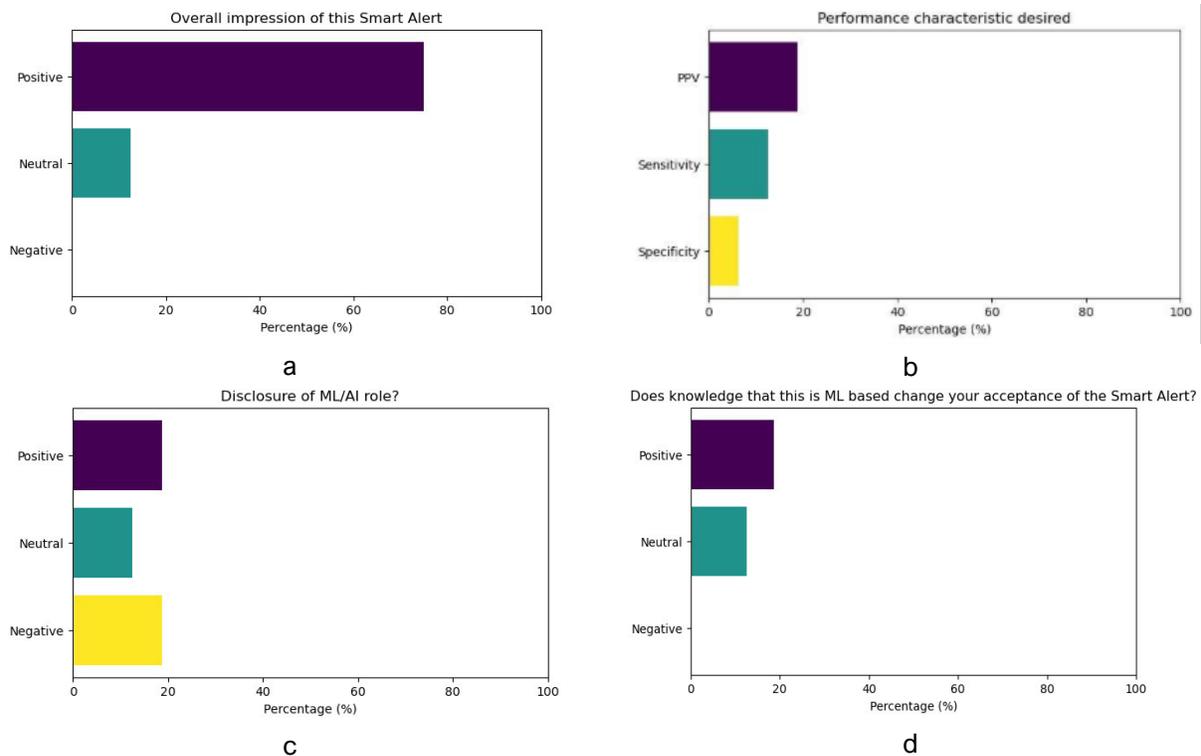

**Figure 3: Clinician perspectives of the alert in pre-deployment interviews.** 18 trainees, attending physicians, and advanced practice providers from Internal Medicine/Hospital Medicine, Neurology, Critical Care, General Surgery, Orthopedic Surgery, Hematology/Oncology, Urology, and Otolaryngology were interviewed. **a:** Users overall felt positive about the alert. **b:** PPV was the most desired test performance characteristic users wanted to know about the underlying ML algorithm. **c:** Users were divided on whether the ML component needed to be explicitly disclosed in the SmartAlert interface. **d:** Knowing the

SmartAlert was driven by ML made positive or no impact on user acceptance of the alert recommendations.

**Pilot Approval and 90-day Review**

The alert interface, workflow triggers, and results from the silent prospective evaluation were reviewed in detail at a follow-up CDS Committee meeting (Figure 2). The CDS committee approved pilot deployment of the CBC utilization SmartAlert in eight inpatient units to be followed by a 90-day review of pilot results. In addition to this routine review, the committee also requested a FURM (Fair, Useful, and Reliable AI Models) assessment (Figure 2) given the novel nature of the custom ML model powering the alert. At the 90-day CDS committee review, we reported pilot results (see 'Results') and the committee approved expansion of the alert, conditional on final FURM recommendations.

**FURM (Fair, Useful, and Reliable AI Model) Assessment**

Following our institution's governance process on clinical ML/AI tools, we submitted the CBC utilization SmartAlert for FURM assessment by the Data Science team at Stanford Health Care. The FURM framework, described in detail in Callahan et al[19], is a standardized process involving the project team, relevant stakeholders, and experts in AI model evaluation to assess the potential utility and impact of ML/AI-powered clinical applications. Their findings were reviewed by our institution's Data Science Executive Committee, including leaders from information technology, clinical informatics, and clinical operations, for final approval of enterprise deployment of the CBC utilization SmartAlert.

**Pilot**

Following CDS committee approval of a limited deployment, a pilot of the CBC utilization SmartAlert was conducted from August 15, 2024, to March 15, 2025, in eight acute care inpatient units across two hospitals. Each patient admission was randomized using a custom data field (Epic SmartDataElement) set to either 1 or 2 with equal probability when a patient's chart is first opened during an inpatient encounter. Encounters assigned 1 had the CDS alert display to clinicians when workflow criteria were satisfied ("treatment"), and encounters assigned 2 had the CDS alert hidden ("control") (Figure 4). Primary bone marrow transplant and hematology patient admissions were excluded.

Because the SmartAlert intervention includes both discontinuation and the option to space CBCs to every other day, we defined the primary outcome as the mean number of CBC results within 52 hours of the alert trigger. This window accommodates alternate-day schedules while allowing for a few hours' variation in routine phlebotomy timing. Secondary outcomes were mean number of CBC results within 28 hours of the alert trigger, ICU transfers within 52 hours of the alert trigger, encounter-level ICU transfer rate, length of stay, readmission rate, and encounter-level mortality. Poisson regression tests were used to compare the mean number of CBC results in the timeframes of interest following the alert. The Mann-Whitney U-test was used to compare lengths of stay, and the Fisher's exact test for encounter-level ICU transfers, readmission rates, and mortality.

As part of post-deployment monitoring, all clinicians who interacted with the alert in live clinical settings within the first two months were invited to participate in a qualitative interview. Based on 9 clinicians' feedback, we adjusted the alert criteria to exclude patients with a procedure or blood transfusion in the past 48 hours, or who were on therapeutic intravenous heparin.

This evaluation was deemed Quality Improvement and given a non-Human Subjects Research determination from the Stanford University Institutional Review Board.

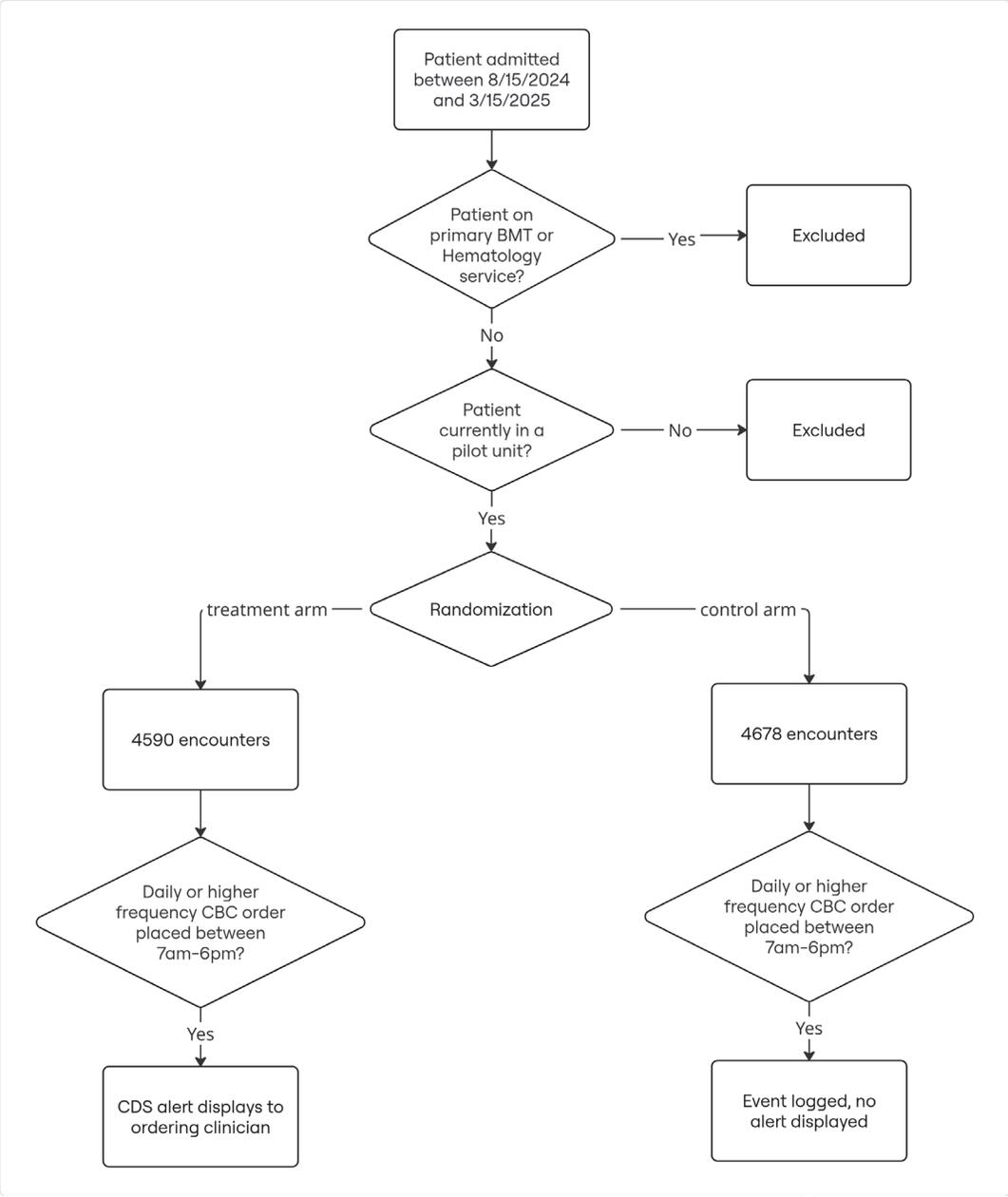

**Figure 4: Randomized controlled pilot study flow diagram.** The CBC utilization SmartAlert was piloted in a randomized controlled fashion from August 15, 2024, to March 15, 2025, in eight acute care inpatient units across two hospitals.

## Results

Over the 7-month pilot, 4,590 eligible encounters were randomized to the treatment arm and 4,678 to the control arm. Patient and encounter features between groups were similar (Table 1). 486 SmartAlerts were displayed to clinicians in the treatment arm and 460 were silently triggered in the control arm. The number of CBC results within 52 hours of the SmartAlert in the treatment arm was 1.54 compared to 1.82 in the control arm (p < 0.01), representing a 15.4% relative reduction. A significant reduction in CBC results was also seen 28 hours following an alert (Table 1). No significant differences were detected in rate of ICU transfers in 52 hours following an alert or in length of stay, readmission rate, or mortality (Table 1).

| Feature | Treatment | Control | P-values |
|---|---|---|---|
| **Encounters** | 4592 | 4678 | |
| **Unique patients** | 4213 | 4261 | |
| **Alerts** | 486 (displayed) | 460 (triggered) | |
| **Median Age [IQR]** | 69 [56-80] | 70 [58-80] | |
| **% Female** | 51 | 50 | |
| **Race (%)** | | | |
|     White | 56 | 54 | |
|     Black or African American | 5 | 6 | |
|     Asian | 16 | 16 | |
|     Pacific Islander | 1 | 1 | |
|     Native American | <1 | <1 | |
|     Other | 19 | 19 | |
|     Unknown | 2 | 2 | |
| **% ICU on admission (Encounter level)** | 0.94 | 1.0 | |
| **Mean number of CBC results within 52 hrs of alert** | 1.54 | 1.82 | < 0.01 |
| **Mean number of CBC results within 28 hrs of alert** | 0.94 | 1.05 | 0.02 |
| **Rate of ICU transfer within 52 hrs of alert (%)** | 0.0 | 0.65 | 0.11 |
| **Length of stay (hours)** | 118.7 | 112.7 | 0.18 |
| **30-day readmission rate (%)** | 11.8 | 12.8 | 0.15 |
| **Encounter mortality rate (%)** | 0.59 | 0.55 | 0.94 |

**Table 1: Demographics and outcomes of CBC utilization SmartAlert pilot.** Patient and encounter features between groups were similar. A significant reduction in CBC results was seen at 52 and 28 hours following an alert in the intervention group, with no significant differences in safety outcomes.

## Discussion

**Lessons Learned**

A SmartAlert system offers precise, personalized guidance more likely to be acceptable to clinicians in reducing laboratory testing. In a randomized pilot trial, the CBC utilization SmartAlert reduced repetitive CBC testing in hospitalized patients without adverse effects on secondary safety measures. In our hospital system, more than 700,000 inpatient CBCs are performed annually, of which over 30% are repetitive. A

15% relative reduction in repetitive CBCs alone would eliminate 31,500 tests annually, corresponding to approximately $13.3 million in institution-specific charges, or $4.1 million using California's median hospital CBC charge.[20] These results demonstrate the technical feasibility of integrating custom prediction models into EHRs more broadly, highlighting the importance of a collaborative governance and implementation structure to translate technical models into practice. Our health system has established practices to evaluate the impact and ensure the acceptability of classic CDS constructs like OurPracticeAdvisory alerts, which ensures early stakeholder engagement, iterative design for usability, and phased deployment to ensure intended behavior prior to broader deployments. The novel integration of ML-based predictions prompted further governance evaluation through a formalized FURM assessment by our health system's Data Science team, evaluating technical feasibility as well as ethical, financial, and clinical implications.

While SmartAlert's FHIR-enabled architecture allows for direct flow from clinical order entry to patient-specific data retrieval calculation and decision support, in practice we found that data calls could take minutes for each patient, resulting in our decision to pursue batched pre-calculation.

**Limitations**

This work was conducted at a single academic medical center with established governance committees that may not exist in other settings, but the underlying system architecture and governance process can be replicated in any setting without proprietary systems or knowledge. Our pilot evaluation focused on general medical and surgical settings, but additional evaluation would be necessary elsewhere (e.g., intensive care units and hematology wards). While CBCs were targeted in this pilot implementation, the flexibility of the SmartAlert framework can be extended to many other diagnostic tests and processes, including repetitive metabolic/chemistry panels, initial laboratory testing, medication ordering, risk predictions, and diagnostic coding suggestions.

## Conclusion

A machine learning-driven system backed by thoughtful implementation, deliberate evaluation, and systematic governance can provide personalized guidance on inpatient laboratory testing to safely reduce unnecessary repetitive testing, benefitting both patient and health systems.

## Acknowledgments

The authors express their gratitude to the Stanford Health Care and Stanford Health Care Tri-Valley teams, as well as participating clinicians for their contributions. This research was supported by the Singapore National Science Scholarship PhD Scholarship (Jiang), the Stanford Department of Pediatrics Fellow Scholarship Award (Liang), and Stanford University. Dr. Chen has received research funding support in part by:

- NIH/National Institute of Allergy and Infectious Diseases (1R01AI17812101)
- NIH-NCATS-Clinical & Translational Science Award (UM1TR004921)
- Stanford Bio-X Interdisciplinary Initiatives Seed Grants Program (IIP) [R12] [JHC]
- NIH/Center for Undiagnosed Diseases at Stanford (U01 NS134358)
- Stanford Institute for Human-Centered Artificial Intelligence (HAI)
- Stanford RAISE Health Seed Grant (2024)

## Supplemental Materials

S1: Pre-pilot clinician interview guide

Current Practices

> What are your experiences with discontinuing or spacing out standing labs in hospitalized patients?
> What factors do you consider when deciding whether to discontinue or space out standing labs on a patient?
> When in your daily workflow do you think about discontinuing or spacing standing labs for a patient?
> When do you actually discontinue or modify the lab order?
> What tools, systems, or reminders, if any, do you use to remember to address standing lab orders?
> We would like to know what you consider to be a stable lab value in order to stop monitoring. Specifically, what is an acceptable relative change over 24 hours, and what is considered a safe absolute value. We would like to capture those thresholds with this table. Taking the first row as an example: I consider an INR value that drops by 0.2 or less or rises by 0.3 or less in 24 hours to be stable, as long as the value falls between 0.8 and 1.3 the next day. Does that make sense?
> Take as much time as you need now to reflect on your practice and fill out this table. You can ask me questions at any time. You are encouraged to think aloud during this process.

| Lab | Reference Range | Acceptable Decrease to Stop Trend | Acceptable Increase to Stop Trend | Acceptable Minimum to Stop Trend | Acceptable Maximum to Stop Trend |
|---|---|---|---|---|---|
| *INR* | *0.9-1.1* | *-0.2* | *+0.3* | *0.8* | *1.3* |
| WBC | 4-11 | | | | |
| Hgb | 12-16 | | | | |
| Plt | 150-400 | | | | |
| Na | 135-145 | | | | |
| K | 3.5-5.5 | | | | |
| Cr | 0.5-1.0 | | | | |
| Ca | 8.4-10.5 | | | | |
| Mg | 1.6-2.4 | | | | |
| Phos | 2.5-4.5 | | | | |
| AST | 10-35 | | | | |
| ALT | 10-35 | | | | |
| Tbili | < 1.2 | | | | |

| Alk Phos | 35-105 | | | | |

  Any thoughts you have while filling out the table?

In this last portion of the interview, I will share my screen to show you an example design of an Epic best practice alert (BPA) that advises providers on when standing lab testing may be low yield and provides options to change the lab orders.

  What do you think of this alert?
  What can be improved about this alert?
  How do you feel about the information provided on this alert?
  How many recent results would you want to see?
  Are there any items that you feel are irrelevant or extraneous?
  Are there any items that you feel are missing?
  What do you think the default action pre-selected for the provider should be?
  What do you think of the available override reasons?
  Overall is there too much information to process while working clinically?
  Whether this alert appears will be based on a machine learning algorithm predicting whether the patient's next lab result will be stable. How do you feel about this alert now knowing about the machine learning aspect?
  The BPA never explicitly states that the predictions were generated by an ML algorithm, how do you feel about that omission?
  What information, if any, do you need to trust this algorithm's predictions?
  Algorithm's prediction threshold? Probability of this particular prediction?
  How was the algorithm trained? What kind of model?
  Publication? Validation?
  Algorithm's performance characteristics (e.g. sensitivity/specificity, PPV/NPV)?
  Algorithm's prediction of next lab value or confidence interval?
  This alert is not yet integrated into the clinical workflow. Where, if at all, would you integrate this alert into your current workflow? For example:
  When you click "re-order" on a standing daily lab?
  When you are reviewing lab results in AM?
  When you open Orders activity between 10am-5pm?
  How often should the alert pop up?
  Should the snooze period be different depending on which override reason you chose? If so, what periods for each reason?

S2: Clinician-defined consensus stability thresholds for common inpatient lab components based on focus group of 18 clinicians.

Participating providers included 4 Internal Medicine/Hospital Medicine, 4 Neurology, 3 Critical Care, 3 General Surgery, 1 Orthopedic Surgery, 1 Hematology/Oncology, 1 Urology, and 1 Otolaryngology.

| Lab Component | Reference Range | Acceptable Decr to Stop Trend (mean +/- std) | Acceptable Incr to Stop Trend (mean +/- std) | Acceptable Min to Stop Trend (mean +/- std) | Acceptable Max to Stop Trend (mean +/- std) |
|---|---|---|---|---|---|
| **WBC** | 4-11 | -2.7 +/- 1.2 | +1.8 +/- 1.0 | 4.6 +/- 2.5 | 11.6 +/- 1.1 |
| **Hgb** | 12-16 | -0.99 +/- 0.68 | +1.9 +/- 1.0 | 9.5 +/- 1.6 | 16.4 +/- 0.98 |
| **Plt** | 150-400 | -36.5 +/- 29.0 | +65.6 +/- 36.8 | 124.7 +/- 52.5 | 496.1 +/- 152.9 |
| **Na** | 135-145 | -3.4 +/- 1.7 | +3.7 +/- 1.7 | 131.6 +/- 2.4 | 146.2 +/- 1.8 |
| **K** | 3.5-5.5 | -0.49 +/- 0.41 | +0.63 +/- 0.62 | 3.4 +/- 0.29 | 5.1 +/- 0.31 |
| **Cr** | 0.5-1.0 | -0.39 +/- 0.26 | +0.28 +/- 0.11 | 0.37 +/- 0.26 | 2.2 +/- 2.6 |
| **Ca** | 8.4-10.5 | -0.74 +/- 0.31 | +0.85 +/- 0.31 | 8.1 +/- 0.30 | 10.6 +/- 0.41 |
| **Mg** | 1.6-2.4 | -0.43 +/- 0.28 | +0.53 +/- 0.44 | 1.7 +/- 0.30 | 2.8 +/- 0.41 |
| **Phos** | 2.5-4.5 | -0.47 +/- 0.22 | +0.86 +/- 0.71 | 2.5 +/- 0.24 | 4.7 +/- 0.38 |
| **AST** | 10-35 | -19.5 +/- 15.1 | +19.2 +/- 21.4 | 8.1 +/- 8.7 | 106.3 +/- 223.9 |
| **ALT** | 10-35 | -19.5 +/- 15.1 | +19.2 +/- 21.4 | 8.1 +/- 8.7 | 106.3 +/- 223.9 |
| **Tbili** | < 1.2 | -0.47 +/- 0.35 | +0.36 +/- 0.21 | 0.34 +/- 0.44 | 2.0 +/- 2.0 |
| **Alk Phos** | 35-105 | -25.9 +/- 23.7 | +25.3 +/- 21.9 | 23.1 +/- 17.6 | 155.8 +/- 76.1 |

S3: Pilot deployment user interview guide

Current Practices

What are your experiences with discontinuing or spacing out standing labs in hospitalized patients?
What factors do you consider when deciding whether to discontinue or space out standing labs on a patient?
When in your daily workflow do you think about discontinuing or spacing standing labs for a patient?
When do you actually discontinue or modify the lab order?
What tools, systems, or reminders, if any, do you use to remember to address standing lab orders?
We would like to know what you consider to be a stable lab value in order to stop monitoring. Specifically, what is an acceptable relative change over 24 hours, and what is considered a safe absolute value. We would like to capture those thresholds with this table. Taking the first row as an example: I consider an INR value that drops by 0.2 or less or rises by 0.3 or less in 24 hours to be stable, as long as the value falls between 0.8 and 1.3 the next day. Does that make sense?
Take as much time as you need now to reflect on your practice and fill out this table. You can ask me questions at any time. You are encouraged to think aloud during this process.

| Lab | Reference Range | Acceptable Decrease to Stop Trend | Acceptable Increase to Stop Trend | Acceptable Minimum to Stop Trend | Acceptable Maximum to Stop Trend |
|---|---|---|---|---|---|
| *INR* | *0.9-1.1* | *-0.2* | *+0.3* | *0.8* | *1.3* |
| WBC | 4-11 | | | | |
| Hgb | 12-16 | | | | |
| Plt | 150-400 | | | | |
| Na | 135-145 | | | | |
| K | 3.5-5.5 | | | | |
| Cr | 0.5-1.0 | | | | |
| Ca | 8.4-10.5 | | | | |
| Mg | 1.6-2.4 | | | | |
| Phos | 2.5-4.5 | | | | |
| AST | 10-35 | | | | |
| ALT | 10-35 | | | | |
| Tbili | < 1.2 | | | | |

| Alk Phos | 35-105 | | | | |
|---|---|---|---|---|---|

Any thoughts you have while filling out the table?

Pilot BPA Review

In this last portion of the interview, we want to hear about your experiences with the lab utilization BPA that is being piloted. This screen shows the alert in Epic.

- Do you recall seeing this alert in your recent work? What was your experience with it?
- How did you feel about this alert?
- In instances where you chose to alter the labs as suggested, if any, what were the reasons?
- In instances where you chose NOT to alter the labs, if any, what were the reasons?
- Did you click into the "Learn more" link to learn more about stability calculation?
- What can be improved about this alert?
- How did you feel about the content of this alert?
- Amount of information in the alert? Is there too much to process while working clinically?
- Irrelevant or extraneous information?
- Missing information?
- How many recent results would you want to see?
- What do you think of the default action being set to discontinue the current lab?
- What did you think of the available alternative CBC choices?
- What did you think of the available override reasons?
- How did you feel about the timing of when this alert came up during your workflow?
- When in your workflow would you like this alert to pop up?
- How did you feel about the frequency with which this alert came up?
- Should it differ based on which acknowledge reason was chosen?
- How often would you like this alert to pop up?
- Whether this alert appears is based on a machine learning algorithm predicting whether the patient's next lab result will be stable. How do you feel about this alert now knowing about the machine learning aspect?
- The BPA never explicitly states that the predictions were generated by an ML algorithm, how do you feel about that omission?
- What additional information, if any, would you need to trust this algorithm's predictions?
- Algorithm's prediction threshold? Probability of this particular prediction?
- How was the algorithm trained? What kind of model?
- Publication? Validation?
- Algorithm's performance characteristics (e.g. sensitivity/specificity, PPV/NPV)?
- Algorithm's prediction of next lab value or confidence interval?

S4: Evolution of interface design of the lab utilization SmartAlert a: Initial alert mockup presented at the August 2023 CDS Committee at concept approval stage.

b: Updated mockup shown to clinicians in co-design interviews between October 2023 and April 2024. Provider feedback highlighted preference for less text, 3 prior results, and the clear prediction probability (PPV).

c: EHR test build version 1. Prior results display vertically.

d: EHR test build version 2. Prior results display horizontally.

e: EHR test build version 3. Time of prediction generation provided. Weekly lab frequency added as alternative. Explanatory text and link to algorithm information adjusted.

f: EHR test build version 4. Abnormal values highlighted in red. Additional lab frequency alternatives provided as an order panel action.